\documentclass[10pt,twocolumn,letterpaper]{article}

\usepackage{iccv}
\usepackage{times}
\usepackage{epsfig}
\usepackage{graphicx}
\usepackage{amsmath}
\usepackage{amssymb}
\usepackage{wrapfig}
\usepackage{float,wrapfig}
\usepackage{color,xcolor}
\usepackage{epsfig}
\usepackage{graphicx}

\usepackage{adjustbox} 
\usepackage{array}
\usepackage{booktabs}
\usepackage{colortbl}
\usepackage{float,wrapfig}
\usepackage{hhline}
\usepackage{multirow}
\usepackage{subcaption} 

\usepackage{amsmath,amsfonts,amsthm,amssymb}
\usepackage{bm}
\usepackage{nicefrac}
\usepackage{microtype}

\usepackage{changepage}
\usepackage{extramarks}
\usepackage{fancyhdr}
\usepackage{lastpage}
\usepackage{setspace}
\usepackage{soul}
\usepackage{xspace}
\usepackage{placeins}
\usepackage{url}

\usepackage{algorithm, algorithmic}
\usepackage{enumerate}

\iccvfinalcopy 

\ificcvfinal\fi
\begin{document}

\title{High-Resolution Shape Completion Using Deep Neural Networks for\\Global Structure and Local Geometry Inference}

\author{Xiaoguang Han$^{1,}$$^{*}$  \quad Zhen Li$^{1,}$$^{*}$   \quad Haibin Huang$^{2}$   \quad Evangelos Kalogerakis$^{2}$ \quad Yizhou Yu$^{1}$\\$^{1}$The University of Hong Kong \qquad \qquad $^{2}$University of Massachusetts, Amherst}


\twocolumn[{
\maketitle
\begin{center}
\centering
\includegraphics[width=0.98\textwidth]{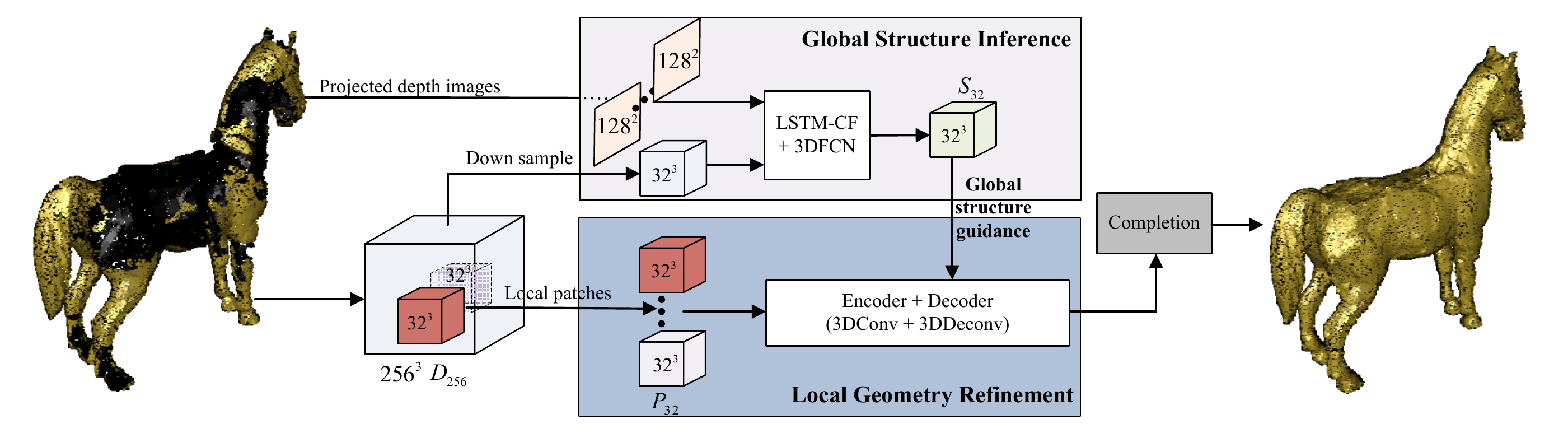}
\captionof{figure}{Pipeline of our high-resolution shape completion method. Given a 3D shape with large missing regions, our method outputs a complete shape through global structure inference and local geometry refinement. Our architecture consists of two jointly trained sub-networks: one network predicts the global structure of the shape while the other locally generates the repaired surface under the guidance of the first network.}
\label{fig:introduction}
\end{center}
 }]

\begin{abstract}
\vspace{-5mm}
We propose a data-driven method for recovering missing parts of 3D shapes. Our method is based on a new deep learning architecture consisting of two sub-networks: a global structure inference network and a local geometry refinement network. The global structure inference network incorporates a long short-term memorized context fusion module (LSTM-CF) that infers the global structure of the shape based on multi-view depth information provided as part of the input. It also includes a 3D fully convolutional (3DFCN) module that further enriches the global structure representation according to volumetric information in the input. Under the guidance of the global structure network, the local geometry refinement network takes as input local 3D patches around missing regions, and progressively produces a high-resolution, complete surface through a volumetric encoder-decoder architecture. Our method jointly trains the global structure inference and local geometry refinement networks in an end-to-end manner. We perform qualitative and quantitative evaluations on six object categories, demonstrating that our method outperforms existing state-of-the-art work on shape completion.

\renewcommand{\thefootnote}{*}
\footnotetext{equal contribution}

\end{abstract}

\section{Introduction}
Inferring geometric information for missing regions of 3D shapes is a fundamental problem in the fields of computer vision, graphics and robotics. With the increasing availability of consumer depth cameras and geometry acquisition devices, robust reconstruction of complete 3D shapes from noisy, partial geometric data remains a challenging problem. In particular, a significant complication is the existence of large missing regions in the acquired 3D data due to occlusions, reflective material properties, and insufficient lighting conditions. Traditional geometry-based methods, such as Poisson surface reconstruction (\cite{kazhdan2006poisson}), are only able to handle relatively small gaps in the acquired 3D data. Unfortunately, these methods often fail to repair large missing regions. Learning-based approaches are more suitable for this task because of their ability to learn powerful 3D shape priors from large online 3D model collections (e.g., ShapeNet, Trimble Warehouse) for repairing such missing regions.

In the past, volumetric convolutional networks have been utilized ~\cite{dai2016complete} to learn a mapping from an incomplete 3D shape to a complete one, where both the input and output shapes are represented with voxel grids. Due to the high computational cost and memory requirement of three-dimensional convolutions, the resolution of the voxel grids used in these methods is severely limited ($32^{3}$ in most cases). Such a coarse representation gives rise to loss of surface details as well as low-resolution, implausible outputs. Post-processing methods, such as volumetric patch synthesis, could be applied to refine the shape, yet producing a high-quality shape still remains challenging as synthesis starts from a low-resolution intermediate result in the first place.

In this paper, we propose a deep learning framework pursuing high-resolution shape completion through joint inference of global structure and local geometry. Specifically, we train a global structure inference network including a 3D fully convolutional (3DFCN) module and a view-based long short-term memorized context fusion module (LSTM-CF). The representation generated from these modules encodes the inferred global structure of the shape that needs to be repaired. Under the guidance of the global structure inference network, a 3D encoder-decoder network reconstructs and fills missing surface regions. This second network operates at the local patch level so that it can synthesize detailed geometry. Our method jointly trains these two sub-networks so that it is not only able to infer an overall shape structure but also refine local geometric details on the basis of the recovered structure and context.

Our method utilizes the trained deep model to convert an incomplete point cloud into a complete 3D shape. The missing regions are progressively reconstructed patch by patch starting from their boundaries. Experimental results demonstrate that our method is capable of performing high-quality shape completion. Qualitative and quantitative evaluations also show that our algorithm outperforms existing state-of-the-art methods.


In summary, this paper has the following contributions:
\begin{itemize}
   \item{A novel global structure inference network based on a 3D FCN and LSTM. It is able to map an incomplete input shape to a representation encoding a complete global structure.}

    \item{A novel patch-level 3D CNN for local geometry refinement under the guidance of our global structure inference network. Our patch-level network is able to perform detailed surface synthesis from the starting point of a low-resolution voxel representation.}

    \item{A pipeline for jointly training the global structure and local geometry inference networks in an end-to-end manner.}
\end{itemize}

\section{Related work}
There exist a large body of work on shape reconstruction from an incomplete point cloud. A detailed survey can be referred to~\cite{berger2014state}.

\paragraph{Geometric approaches.} By assuming local surface or volumetric smoothness, a number of geometry-based methods (\cite{kazhdan2006poisson}~\cite{tagliasacchi2011vase}~\cite{wu2015deep}) can successfully recover the underlying surface in the case of small gaps. To fill larger missing regions, some methods employ hand-designed heuristics for particular shape categories. For example, Schnabel {\it{et al.}}~\cite{schnabel2009completion} developed an approach to reconstruct CAD objects from incomplete point clouds, under the assumption that the shape is composed of many primitives (e.g., planes, cylinders, cones etc.). Li {\it{et al.}}~\cite{li2011globfit} further considered geometric relationships (eg. orientation, placement, equality, etc.) between primitives in the reconstruction procedure. For  objects with arterial-like structures, Li {\it{et al.}}~\cite{li2010analysis} proposed snake deformable models and successfully recovered the topology and geometry simultaneously from a noisy and incomplete input. Based on the observation that urban objects are usually made of non-local repetitions, many methods (~\cite{pauly2008discovering}~\cite{zheng2010non}) attempt to discover symmetries from input data and use them to complete the unknown geometry. Harary {\it{et al.}}~\cite{harary2014context} also utilizes self-similarities to recover shape surfaces. Compared to these techniques, we propose a learning-based approach that learns a generic 3D\ shape prior for  reconstruction without resorting to hand-designed heuristics or strict geometric assumptions.

\paragraph{Template-based approaches.} Another commonly used strategy is to resort to deformable templates or nearest-neighbors to reconstruct an input shape. One simple approach is to retrieve the most similar shape from a database and use it as a template that can be deformed to fit the input raw data (~\cite{pauly2005example}~\cite{rock2015completing}). These template-based approaches usually require user interaction to specify sparse correspondences (~\cite{pauly2005example}) or result in wrong structure (~\cite{rock2015completing}) especially for complex input. These approaches can also fail when the input does not match well with the template, which  often happens due to the limited capacity of the shape database. To address this issue, some recent works (~\cite{shen2012structure}~\cite{sung2015data}) involved the concept of part assembly. They can successfully recover the underlying structure of a partial scan by solving a combinatorial optimization problem that aims to find the best part candidates from a database as well as their combination. However, these methods also have a number of limitations. First, each shape in the database needs to be accurately segmented and labeled. Second, for inputs with complicated structure, these methods may fail to  find the global optimum due to the large solution space. Lastly, even if  coarse structure is well recovered, obtaining the exact underlying geometry for missing regions remains challenging especially when the input geometry does not match well any shape parts in the database.

\paragraph{Deep learning-based methods.} Recently, 3D convolutional networks have been proposed for shape completion. Wu {\it{et al.}}~\cite{wu20153d} learn a probability distribution over binary variables representing voxel occupancy in a 3D grid based on Convolutional Deep Belief Networks (CDBNs). CDBNs are generative models that can also be applied for shape completion. Nguyen {\it{et al.}}~\cite{nguyen2016field} combines CDBNs and Markov Random Fields (MRFs) to   formulate shape completion as a Maximum a Posteriori (MAP) inference problem. More recent methods employ encoder-decoder networks that are trained~end-to-end for shape reconstruction \cite{sharma16arxiv,varley2016shape}. However, all these techniques operate on low-resolution grids ($30^{3}$ voxels to represent global shape) due to the high computational cost of convolution in three dimensions. The recent work of Dai {\it{et al.}}~\cite{dai2016complete} is most related to ours. It proposes a 3D Encoder-Predictor Network (EPN) to infer a coarse shape with  complete structure, which is then further refined through nearest-neighbor-based volumetric patch synthesis. Our method also learns a global shape structure  model. However, in contrast to Dai {\it{et al.}}, we also learn a \emph{local} encoder-predictor network to perform patch-level surface inference.\ Our network produces a more detailed output in a much higher resolution ($256^{3}$ grid) by
processing local shape patches through this network ($32^{3}$ patches cropped from the $256^{3}$ grid).  Local surface inference is performed under the guidance of our global structure  network that captures the necessary contextual information to achieve a globally consistent, and at the same time, high-resolution reconstruction.

\section{Overview}
Given a partial scan or an incomplete 3D object as  input, our method aims to generate a complete object as  output. At a high level, our pipeline is similar to PatchMatch-based image completion~\cite{barnes2009patchmatch}. Starting from the boundary of missing regions, our method iteratively extends the surface into these regions, and at the same time updates their boundary for further completion until these missing regions are filled. To infer new geometry along the boundary, instead of retrieving the best matching patch from a large database as in \cite{harary2014context}, we designed a local surface inference model based on volumetric encoder-decoder networks.

The overall pipeline of our method is shown in Figure~\ref{fig:introduction}. Our shape completion is performed patch-by-patch. At first, the input point cloud is voxelized in a $256^{3}$ grid, and then  $32^{3}$ patches are extracted along the boundary of missing regions. Our local surface  inference network maps the volumetric distance field of a surface, potentially with missing regions, to an implicit representation (0 means inside while 1 means outside) of a complete shape (Section~\ref{sec:local_net}). The distance field can then be extracted with Marching Cubes~\cite{lorensen1987marching}. To improve the global consistency of local predictions during shape completion, another global structure inference network is designed to infer  complete global shape structure and guide the local geometry refinement network. 
Our global structure inference network generates a  $32^{3}$  shape representation, capturing its overall coarse structure, using both view-based and volumetric deep neural networks. To make use of high-resolution shape information, depth images generated over the six faces of the bounding cube are considered as one type of input data (view-based input). Six 2D feature representations of these depth images are extracted through six parallel streams of 2D convolutional and LSTM recurrent layers. These 2D feature maps are assembled into a $32^{3}$ feature representation, which is fused with another volumetric-based feature representation extracted through 3D convolutional layers operating on volumetric input. The resulting fused representation is used for final voxel-wise prediction.  Both our global and local network are trained jointly (Section~\ref{sec:architecture}).

\section{Network Architecture}
\label{sec:architecture}
The incomplete point cloud is represented as a $256^{3}$ volumetric distance field (Section~\ref{sec:training}), denoted as $D_{256}$. Our deep neural network is composed of two sub-networks. One infers underlying global structure from a down-sampled version ($32^{3}$) of $D_{256}$. The down-sampled field is denoted as $D_{32}$ and the inferred result is denoted as $S_{32}$. Another sub-network infers high-resolution local geometry within $32^{3}$ volumetric patches (denoted as $P_{32}$) cropped from $D_{256}$.

\subsection{Global Structure Inference}
\begin{figure}[t]
\centering
\includegraphics[width=0.46\textwidth]{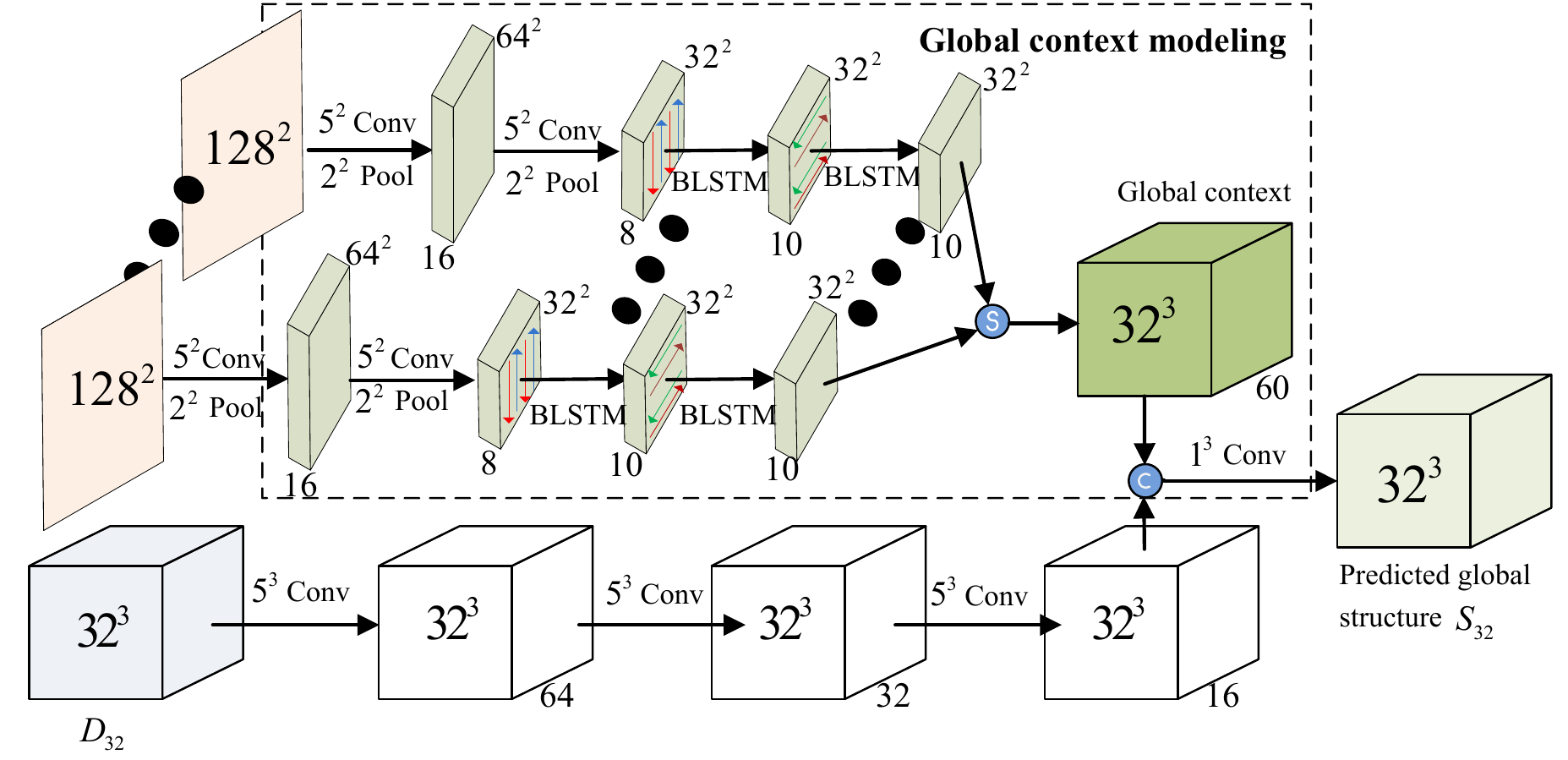}
\caption{The inputs for global structure inference consist of projected depth images with size $128^2$ and down-sampling voxelized point cloud $D_{32}$ with resolution $32^3$. ``S'' stands for the feature stack operation. ``C'' stands for the concatenate operation.}
\label{fig:global}
\end{figure}
Although an incomplete point cloud has missing data, most often it still provides adequate information for recognizing the object category and understanding its global structure (i.e. object parts and their spatial layout). Such categorical and structural information provides a global context that can help resolve ambiguities arising in local shape completion. Therefore, there is a need to automatically infer the global structure of the underlying object given an incomplete point cloud of the object.

To this end, we design a novel deep network for global structure inference. This network takes two sets of input data. Since processing $D_{256}$ would be too time- and memory-consuming, the first set of input is $D_{32}$, the down-sampled distance field of the input point cloud. To compensate for the low resolution of $D_{32}$, the second set of input data consists of six $128^2$ depth images, each obtained as an orthographic depth image of the point cloud over one of the six faces of its bounding box. Inspired by the LSTM-CF model~\cite{li2016lstm} and ReNet~\cite{visin2015renet}, each depth image passes through two convolutional layers, each of which is followed by a max-pooling layer. The feature map from the second pooling layer is further fed into the ReNet, which consists of cascaded vertical and horizontal bidirectional LSTM~(BLSTM) layers for global context modeling,
\begin{align}
\nonumber &h^v_{i,j} = \text{BLSTM}(h^v_{i,j-1}, f_{i,j}),~~~~ \text{for}~~ j= 1, \ldots, 32; \\
&h^h_{i,j} = \text{BLSTM}(h^h_{i-1,j}, h^v_{i,j}),~~~~ \text{for}~~ i= 1, \ldots, 32,
\end{align}

\begin{wrapfigure}{r}{0.25\textwidth}
\includegraphics[width=0.25\textwidth]{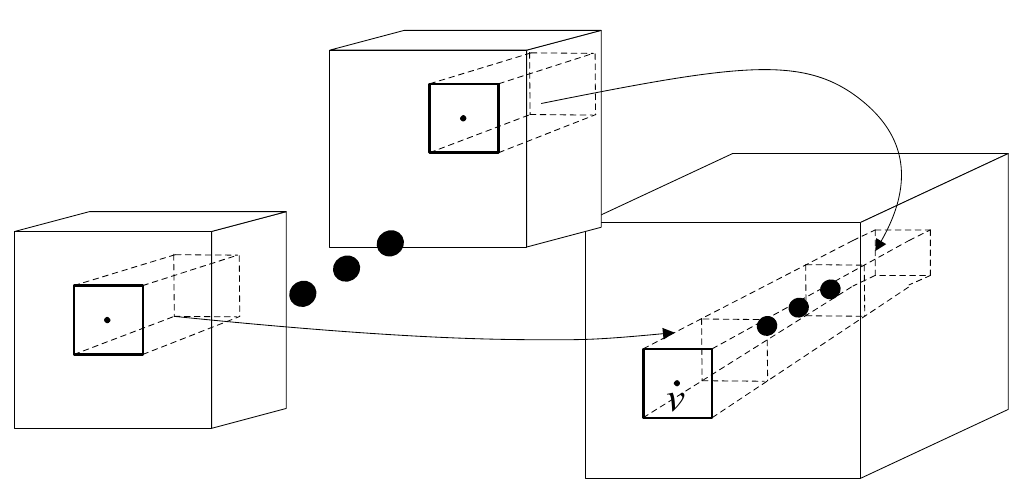}
\caption{3D feature map assembles from six 2D feature maps.}\label{fig:stack}
\end{wrapfigure}
where $f_{i,j}$ is the feature map from the second max pooling layer, $h^v_{i,j}$ and $h^h_{i,j}$ are the output maps of the vertical and horizontal BLSTMs, respectively. Afterwards, 2D output maps from the six horizontal BLSTMs are assembled together into a 3D feature map with size $32^3$ as follows. A voxel $v$ in the 3D feature map is first projected onto the six faces of its bounding box. The projected location on each face is used to look up the feature vector at that location in the corresponding 2D output map. The six retrieved feature vectors from the six 2D maps are concatenated together as the feature for $v$ (in Fig.~\ref{fig:stack} for details). In parallel, $D_{32}$ is fed into three 3D convolutional layers with the same resolution and a 3D feature map with size $32^3$ is obtained after the third layer. Finally, both 3D features maps from the two parallel branches are concatenated and flow into a 3D convolutional layer with $1\times1\times1$ kernels for voxel-wise binary prediction.

It is worth mentioning that the occupancy grids are sparse when used for representing voxelized point clouds. This results in highly uneven distributions of two-class data (inside and outside). For instance, the ratio between the inside and outside voxels for the `chair' category is $25$. Meanwhile, precision and recall both play an importance role in shape completion, especially for inside voxels. To address this problem, we add the AUC loss to the conventional cross-entropy loss for classification ~\cite{calders2007efficient,cortes2003auc}. According to~\cite{calders2007efficient}, the AUC of a predictor $f$ is defined as (here $f$ is the final classification layer with softmax activation illustrated in Fig.~\ref{fig:global}) $\mbox{AUC}(f) = P(f(t_0) < f(t_1) \large \vert t_0 \in D^0, t_1 \in D^1)$, where $D^0, D^1$ are the samples with groundtruth labels 0 and 1, respectively. Its unbiased estimator, i.e. Wilcoxon-Man-Whitney statistics, is $ n_0 n_1 \mbox{AUC}(f) = \sum_{t_0\in D^0} \sum_{t_1 \in D^1} I [ f(t_0) < f(t_1) ] $, where $n_0 = |D^0|, n_1 = |D^1|$, and $I$ is the indicator function. In order to add the noncontinuous AUC loss to the continuous cross-entropy loss and optimize the combined loss through gradient decent, we consider an approximation of the AUC loss by a polynomial with degree $k$~\cite{calders2007efficient}, i.e.
\begin{align*}
n_0 n_1 \text{loss}_{\text{auc}} =\sum_{t_0 \in D^0}\sum_{t_1 \in D^1}\sum_{k=0}^d\sum_{l=0}^k \alpha_{kl} f(t_1)^l f(t_0)^{k-l}\\
\end{align*}
where $\alpha_{kl} = c_k C_k^l (-1)^{k-l}$ is a constant. Thus, our global loss function can be formulated as
\begin{align}
\text{loss}_{\text{global}} = -\frac{1}{N} \sum_i s^*_i \log({s_i}) - \lambda_1 \text{loss}_{\text{auc}},
\end{align}
where $s_i$ stands for the predicted probability of a label, $s^*_i$ stands for a ground-truth label, $N$ is the number of voxels and $\lambda_1$ is a balancing weight between the cross-entropy loss and the AUC loss.

\subsection{Local Geometry Refinement}
\label{sec:local_net}
We further propose a deep neural network for inferring the high-resolution geometry within a local 3D patch $P_{32}$ (each $32^{3}$ patch is a crop from $D_{256}$) along the boundary of missing regions. Instead of a 3D fully convolutional network, we exploit an encoder-decoder network architecture to achieve this goal. This is because local patches are sampled along the boundary of missing regions, the surface inside a local patch usually suffers from a larger portion of missing data (on average 50\%) than the global shape and fully connected layers are better suited for higher-level inference.

As shown in Fig.~\ref{fig:local}, the first part of our network transforms the input patch into the latent space through a series of 3D convolutional and 3D max pooling layers. This encoding part is followed by two fully connected layers. The decoding part then achieves voxel-wise binary predictions with a series of 3D deconvolutions. In comparison to prior network designs ~\cite{dai2016complete,sharma16arxiv,varley2016shape}, our network has a notable difference, which is the incorporation of global structure guidance. Given $S_{32}$ generated by our global structure inference model, for each input patch $P_{32}$ centered at $(x, y, z)$ in $D_{256}$, we use $S_{32}$ as guidance at two different places of the pipeline. First, a $8^{3}$ patch centered at $(x/8, y/8, z/8)$ is cropped from $S_{32}$ and passes through a 3D convolutional layer followed by 3D max pooling. The resulting $4^{3}$ patch is concatenated with the 3D feature map at the end of the encoding part. Second, a $4^{3}$ patch centered at $(x/8, y/8, z/8)$ is cropped from $S_{32}$ and directly concatenated with the $4^{3}$ feature map at the beginning of the decoding part. Similar to the loss function of the global structure inference network, the loss of our local geometry refinement network is defined as $\text{loss}_{\text{local}} = -\frac{1}{M} \sum_i p^*_i \log({p_i})$, where $p_i$ is the predicted probability of a local geometry, $p^*_i$ is a ground-truth label, and $M$ is the number of voxels.

\subsection{Network Training}
\begin{figure}[t]
\centering
\includegraphics[width=0.46\textwidth]{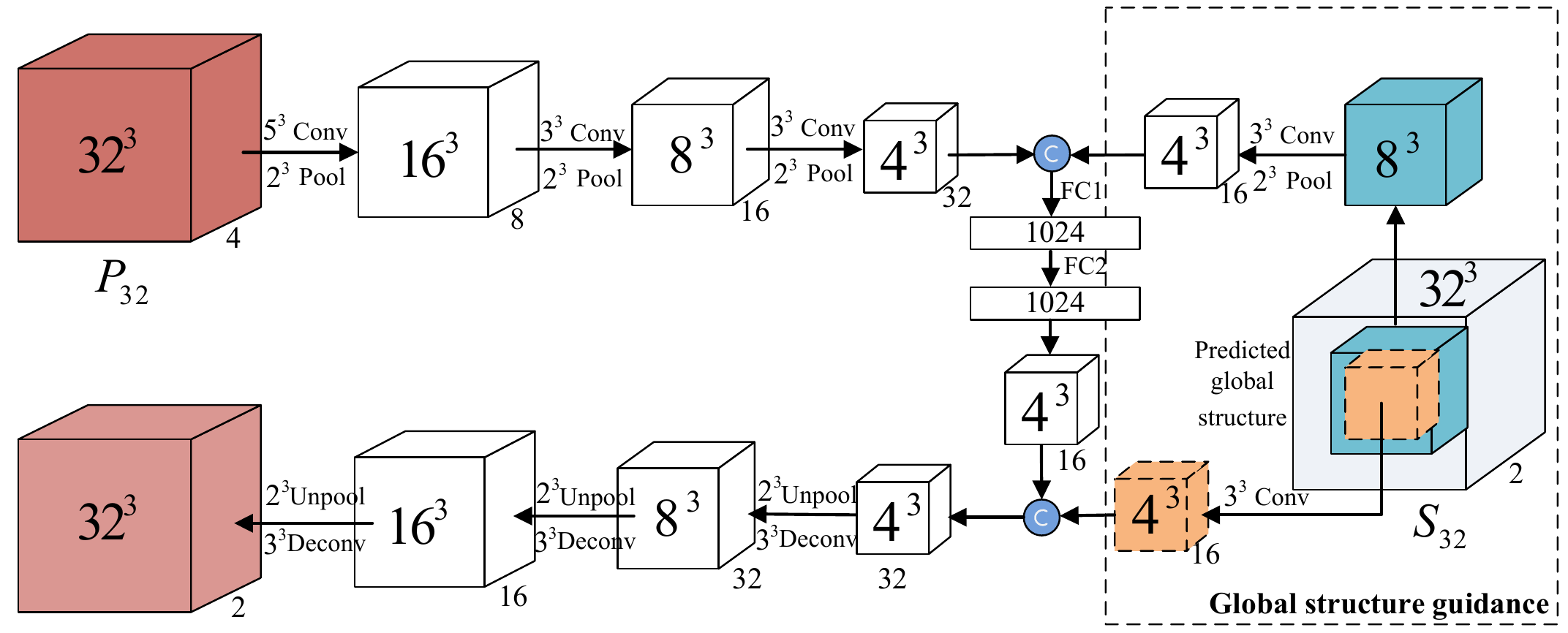}
\caption{The inputs of local surface refinement network consist of local patches $P_{32}$ with size $32^3$ and output $S_{32}$ with size $32^3$ from the global structure inference network as global guidance.}
\label{fig:local}
\end{figure}
Our two deep networks are trained in two phases. In the first phase, the global structure inference network is first trained alone. In the second phase, the local geometry refinement network is trained while the global structure inference network is being fine-tuned.
As illustrated in Fig.\ref{fig:local}, a local patch from $S_{32}$, which is the output from the global structure inference network, flows into the local geometry refinement network as a global guidance, which is vital for shape completion.
On the other side, local patch prediction results can benefit global structure inference as well, e.g., the refined disconnected regions between the side and leg of a chair can give feedback to global structure inference. Thus, due to the interactions between our global and local sub-networks, joint training is performed to improve their performance and robustness. Since the global network has been trained during the first phase, it is further fine-tuned during joint training in the second phase. The loss for such joint training is defined as follows.
\begin{align}
\text{loss} = \text{loss}_{\text{local}} + \lambda_2 \text{loss}_{\text{global}} + \lambda_3 \|\theta\|_2,
\end{align}
where $\lambda_2$ is a balancing weight between the local and global loss functions, $\theta$ is the parameter vector (L$2$ norm is adopted for regression terms in our loss).

\section{Training Data Generation}
\label{sec:training}

Our network has been tested on 6 categories of objects separately. Among them, `chairs', `cars', `guitars', `sofas', and `guns' are from ShapeNet~\cite{chang2015shapenet} and `animals' were collected by ourselves. For each category from ShapeNet, we select a subset of models by removing repeated ones with very similar structures and thin models that cannot be well voxelized on a $32^{3}$ grid. All animal models were also manually aligned in a coordinate system consistent with ShapeNet. To generate training samples, we simulate object scanning using an RGBD camera and create an incomplete point cloud for each model by fusing multiple partial scans with missing regions. On each created incomplete model, we randomly sample $n$ patches along the boundary of missing regions ($n$ is set to 50 for all object categories). The number of created training models for each category are shown in Table ~\ref{tab:sample}. Note that we create multiple scanned models by simulation from each original virtual model. Each point cloud is represented using a volumetric distance field. Note that, to make occupancy grid as dense as possible, different scaling factors are applied to models from different categories before voxelization. This is also the reason why we do not train a single network on all categories.

\begin{table}[ht]
\centering
\caption{Number of training samples}
\label{tab:sample}
\resizebox{0.47\textwidth}{!}
{
\begin{tabular}{|c|c|c|l|l|l|l|}
\hline
Category   & Chair  & Car   & Guitar & \multicolumn{1}{c|}{Gun} & Sofa  & Animal \\ \hline
\# Samples & 1000x3 & 500x5 & 320x5  & 300x5                    & 500x5 & 43x10  \\ \hline
\end{tabular}
}
\end{table}

\paragraph{Virtual Scanning.} We first generate depth maps by placing a virtual camera at 20 distinct viewpoints, which are vertices of a dodecahedron enclosing a 3D model. The camera is oriented towards the centroid of the 3D model. We then randomly select 3-5 viewpoints only to generate partial shapes simulating scans obtained through limited view access. On top of random viewpoint selection, we also randomly add holes and noise to the depth maps to simulate large occlusions or missing data due to specularities and scanner noise. The method in \cite{van2012seeds} is adopted to create holes. For each depth map, this method is run multiple times and super-pixels are generated at a different level of granularity each time.
A set of randomly chosen superpixels are removed at each level of granularity to create holes at multiple scales. The resulting depth maps are then backprojected to the virtual model to form an incomplete point cloud. Note that missing shape regions will also exist due to self occlusions (since depth maps cannot cover the entire object surface).
\paragraph{Colored SDF + Binary Surface.} Each point cloud is converted to a signed distance field as in \cite{dai2016complete}. To enhance the border between points with positive and negative distances, we employ colored SDF (CSDF), which maps negative distances (inside) to colors between cyan and blue and positive distances to colors between yellow and red. As a distance field makes missing parts less prominent, we also take the binary surface (i.e. occupancy grid of input points) as an additional input channel, denoted as BSurf. Thus, a point cloud is converted to a volumetric grid with four channels.

\paragraph{Projected Depth Images.} Our global network takes 6 depth images as the second set of input data. These depth images are orthographic views of the point cloud from the 6 faces of the bounding cube. These depth images are enhanced with jet color mapping~\cite{eitel2015multimodal}. 

\paragraph{Patch Sampling.} In general, there would be a large number of patches with similar local geometry if they were chosen randomly (for example, several chair patches would originate from flat or cylindrical surfaces). To avoid class imbalance and increase the diversity of training samples, we perform clustering on all sampled patches and only the cluster centers are chosen as training patches. Here we only use BSurf as the feature during patch clustering.

\section{Shape Completion}
During testing, given an incomplete point cloud $P$, as the first step, we apply our global structure inference network to generate a complete but coarse structure. As discussed earlier, starting from the boundary of missing regions, our method iteratively extends the surface into these regions until they are completely filled. In this paper, the method from \cite{bendels2006detecting} is used to detect the boundary of missing regions in a point cloud.

During each iteration, local 3D patches with a fixed size of overlap are chosen to cover all points on the boundary of missing regions. Our local geometry refinement network runs on these patches with the guidance from the inferred global structure to produce a voxel-wise probability map for each patch. The probability at each voxel indicates how likely that voxel belongs to the interior of the object represented by the input point cloud. For voxels covered by multiple overlapping patches, we directly average the corresponding probabilities in these patches. Then we transform the resulting probability map into a signed distance field by simply deducting the probability values by 0.5. Marching Cubes~\cite{lorensen1987marching} is then used to extract a partial mesh from the set of chosen patches and a new point set $Q$ is evenly sampled over the partial mesh. We further remove the points in $Q$ that lie very closely to $P$, and detect new boundary points from the remaining points in $Q$. Such detected boundary points form the new boundary of missing regions. The above steps are performed repeatedly until new boundary points cannot be found. In our experiments, 5 iterations are sufficient in most cases.

\section{Experimental Results}


\begin{figure*}[t]
\centering
\includegraphics[width=0.85\textwidth]{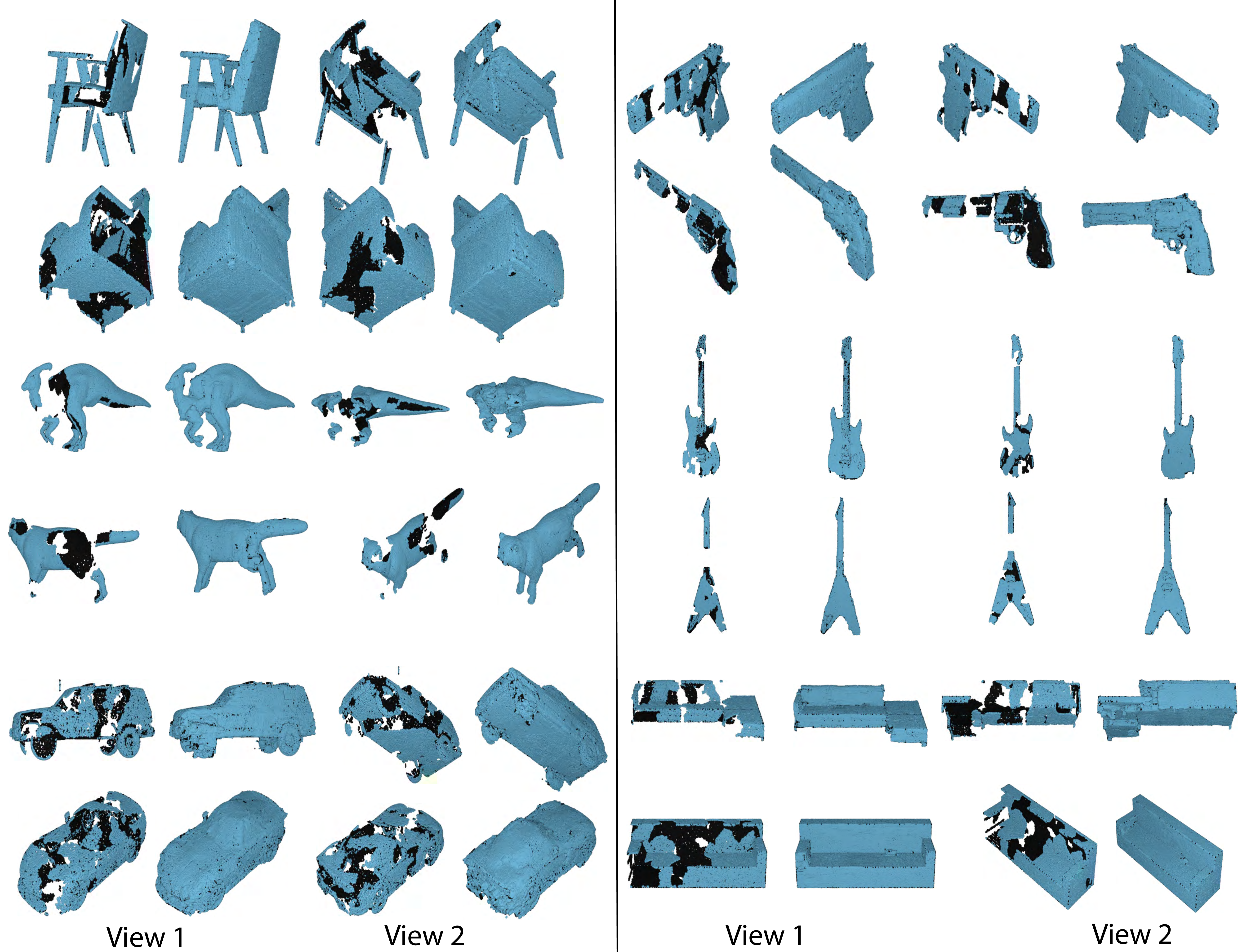}
\caption{Gallery of final results. There are two models per category. For each model, the input and repaired point clouds are shown side by side from two different views.}
\label{fig:result}
\end{figure*}

Fig.~\ref{fig:result} shows a gallery of results. For each object category, two models with different views are chosen. The incomplete point cloud and the repaired result are placed side by side.

\subsection{Implementation}
We jointly train the global and local networks for 20 epochs with an Adam optimizer~\cite{kingma2014adam}. We include one voxelized point cloud, six depth images associated with the point cloud and $50$ local patches sampled from the voxel grid along missing regions in a single mini-batch. The balancing weights are set as follows: $\lambda_1=0.2$ and $\lambda_2=\frac{2}{3}$. The regression weight and learning rate are set to $0.001$ and $0.0001$, respectively. Considering diverse scales in different object categories, we always pre-train our deep network for a specific object category from scratch using $2400$ ($800$ original models with $3$ different virtual scanning per model) chair models. Afterwards, we fine-tune the pre-trained model using training samples from that specific object category. Fine-tuning on each category needs about $4$ hours to converge. On average, it takes 400ms to perform a forward pass through the global structure inference and local geometry refinement networks. Once we have the trained global and local networks, it takes around $60$s to obtain a completed high-resolution shape. The detailed configuration of our global and local networks is illustrated in Figs.~\ref{fig:global} and \ref{fig:local}. The implementation is based on the publicly available Lasagne~\cite{dieleman2015lasagne} library built on the Theano~\cite{BastienTheano2012} platform, and network training is performed on a single NVIDIA GeForce GTX 1080.


\subsection{Comparisons with existing methods}
Let $P_{true}$ be the ground-truth point cloud and $P_{complete}$ be the repaired point cloud. The normalized distance from $P_{complete}$ to $P_{true}$ is used to measure the accuracy of the repaired point cloud. For each point $v \in P_{complete}$, we compute $dist(v, P_{true}) = min\{||v-q||, q\in{P_{true}}\}$. The average of these distances is finally normalized by the maximum shape diameter (denoted as $dm$) across all models in a given category. Following \cite{sung2015data}, we also use ``completeness" to evaluate the quality of shape completion. Completeness records the fraction of points in $P_{true}$ that are within distance $\alpha_{eval}$ of any point in $P_{complete}$. In our setting, $\alpha_{eval}$ is set to $0.001*dm$. The average accuracy and completeness across all models from each object category are reported in Table ~\ref{comparison}. Specifically, we randomly select $200\times3$, $100\times5$, $64\times5$, $60\times5$, $100\times5$ and $8\times10$ samples as the testing set of chairs, cars, guitars, guns, sofas and animals, respectively.

To evaluate the accuracy and efficiency of the proposed method, we perform comparisons against existing state-of-the-art methods. Poisson surface reconstruction~\cite{kazhdan2006poisson} is commonly used to construct 3D models from point clouds. However, it cannot successfully process point clouds with a large percentage of missing data.



\begin{figure}[t]
\centering
\includegraphics[width=0.495\textwidth]{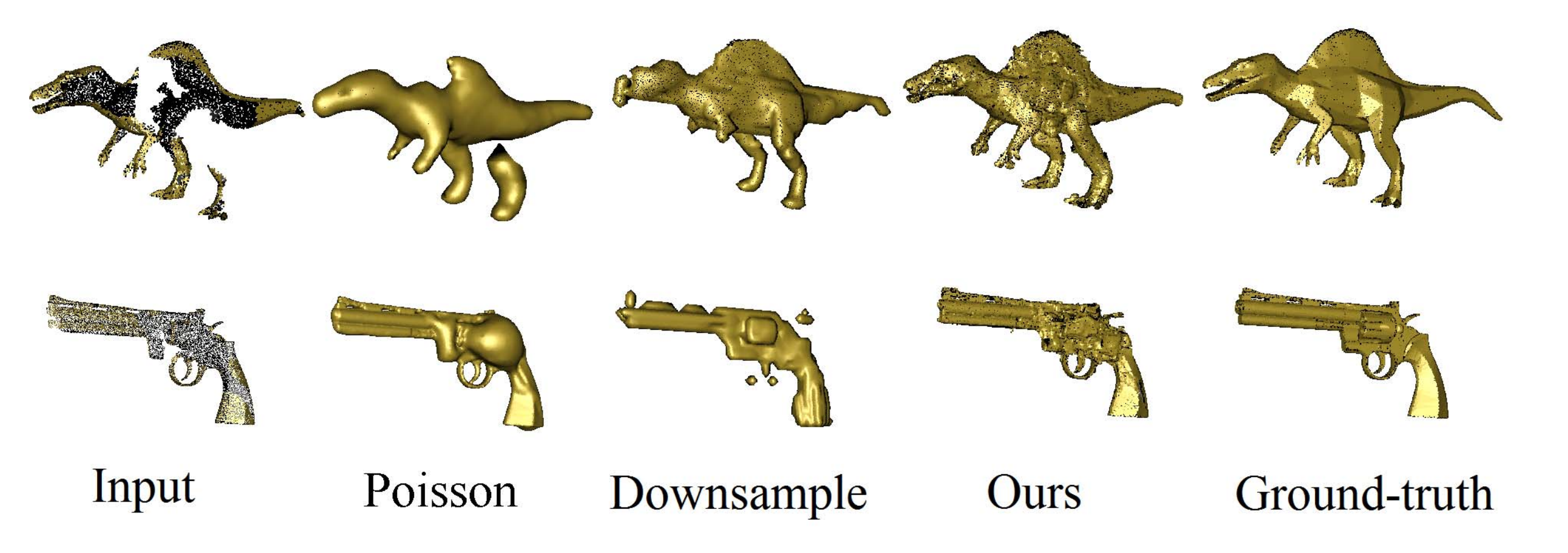}
\caption{Sampled comparison results with other methods.}
\label{fig:comparison}
\end{figure}

Recently, a number of methods ~\cite{sharma16arxiv,varley2016shape,dai2016complete} attempt to perform shape completion at a coarse resolution ($32^{3}$) using 3DCNNs. The completeness and normalized distance of these methods are reported in Table ~\ref{comparison}, where `3D-EPN-unet' stands for the 3D Encoder-Predictor Network with U-net but without 3D classification in ~\cite{dai2016complete}. Note that their model with the 3D classification network is not available. `Vconv-dae' stands for the network from ~\cite{sharma16arxiv} and `Varley {\it{et al.}}' stands for the network from ~\cite{varley2016shape}. For fair comparison, we retrained these networks using our training data and performed the evaluation on our testing data. Evaluation results on our global network are also reported. They demonstrate that our method outperforms all existing $32^3$-level methods even without local geometry inference. To derive an upper bound on the completeness achievable by the methods with $32^{3}$-level outputs, we subsample the distance field of $P_{true}$ at resolution $32^{3}$. We also create a point cloud from that subsampled field (called $P_{downsample}$), which represents a lower bound on the normalized distance that can be achieved by these methods (since these methods introduce additional errors in addition to downsampling). We report the evaluation results on the downsampled point clouds as a baseline for comparison. It can be verified that our results are significantly better than those from existing methods. As an intuitive comparison, Fig.~\ref{fig:comparison} shows the outputs from `Poisson', `$P_{downsample}$' and our method on two sampled models.

The method in \cite{dai2016complete} also proposes a way to achieve high-resolution completion by perfroming 3D patch synthesis as a post-processing step for shape refinement. In comparison to this approach, an important advantage of our method is that our local geometry refinement network directly utilizes the high-resolution information from the input point cloud, which makes the refined results more accurate. Another type of methods~\cite{shen2012structure,sung2015data} can also complete a shape with large missing regions. However, they require a large database with well-segmented objects. As the models in our datasets have not been segmented into parts, we do not conduct comparisons against such methods since a fair comparison seems out of reach.

\begin{table*}[]
\centering
\caption{Performance Comparison. For each category and each method, we show the value of \emph{completeness/normalized dist}. }
\label{comparison}
\resizebox{1.0\textwidth}{!}
{
\begin{tabular}{|c|c|c|c|c|c|c|c|c|}
\hline
Category & Input & Varley {\it{et al.}} & Vcov-dae & 3D-EPN-unet & Our global network & Downsample & Poisson & Our whole network \\ \hline
Chair    & 71.5\%/0       & 35.8\%/0.022           & 47.7\%/0.020      & 58.5\%/0.018         & 70.1\%/0.0108               & 73.32\%/0.0144       & 87.61\%/0.00925  & \textbf{97.25\%/0.00398}   \\ \hline
Car      & 69.1\%/0       & 42.3\%/0.014           & 64.6\%/0.013      & 66.4\%/0.0092        & 81.8\%/0.0081               & 84.35\%/0.00756      & 82.18\%/0.0147   & \textbf{95.88\%/0.00312}   \\ \hline
Guitar   & 85.7\%/0       & 45.8\%/0.013           & 56.6\%/0.011      & 62.9\%/0.0092        & 69.7\%/0.00626              & 72.16\%/0.00675      & 88.4\%/0.148     & \textbf{94.35\%/0.00248}   \\ \hline
Sofa     & 72.31\%/0      & 18.1\%/0.024           & 58.4\%/0.019      & 62.8\%/0.012         & 77.0\%/0.00845              & 85.15\%/0.00615      & 82.78\%/0.027    & \textbf{95.97\%/0.00217}   \\ \hline
Gun      & 62.7\%/0       & 28.5\%/0.0165          & 39.1\%/0.0134     & 49.2\%/0.0132        & 54.3\%/0.0091               & 56.4\%/0.0102        & 77.68\%/0.0114   & \textbf{98.58\%/0.00281}   \\ \hline
Animal   & 69.05\%/0      & 35.6\%/0.0257          & 47.8\%/0.0229     & 56.1\%/0.019         & 82.4\%/0.01137              & 85.14\%/0.0114       & 88.88\%/0.0567   & \textbf{95.53\%/0.00363}   \\ \hline
\end{tabular}
}
\end{table*}


\subsection{Ablation Study}
\begin{figure}[h]
\centering
\includegraphics[width=0.49\textwidth]{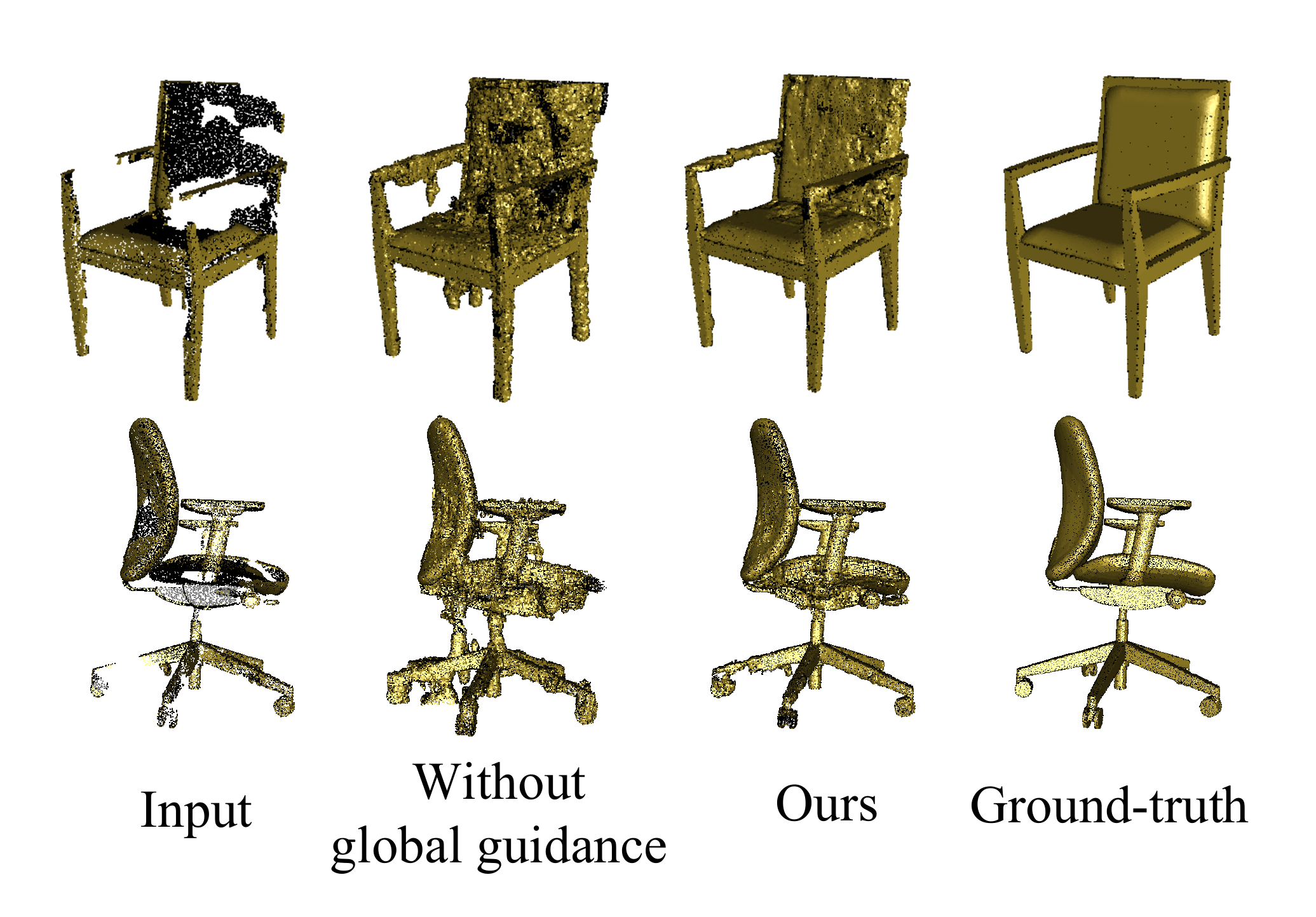}
\caption{Completion results by using our model with and without global guidance.}
\label{fig:withoutglobal}
\end{figure}
To discover the vital elements in the success of our proposed model for shape completion, we conduct an ablation study by removing or replacing individual components in our model trained with $1000\times3$ chair samples, among which $800\times3$ samples form the training set and $200\times3$ samples form the testing set. Specifically, for the global structure inference network, we have tested its performance without the AUC loss, high-resolution depth images, or global context modeling using BLSTM. In addition, we have also tested the global model where the 3DFCN branch only takes CSDF or BSurf as the input to figure out the importance of different input channels. In addition, an encoder-decoder network is used to replace the final 1x1x1 convolutional layer in the global network. For the local geometry refinement network, we have tested its performance by removing the global guidance. The results are presented in Table \ref{table:ablation}. Because of class imbalance, we use the F1-score of the inside labels as the performance measure of the global network. We directly use classification accuracy to evaluate the performance of the local network since class imbalance is not a concern in this case. Note that the ground truth for the global network is defined on a coarse ($32^3$) grid while the ground truth for the local network is defined on a high-resolution ($256^3$) grid.

During this ablation study, we find that CSDF, BSurf and high-resolution depth images are all necessary for global structure inference as the performance drops to $0.90$ if the input to the 3DFCN branch is CSDF only (without the BSurf channel) and the performance drops to $0.877$ if the entire 2D branch taking depth images is eliminated. In addition, the most effective components in our network are BLSTM based context modeling and the AUC loss as the performance drops to $0.896$ and $0.904$, respectively, without either of them. Furthermore, the performance drops to $0.818$ if the final 1x1x1 convolutional layer in the global network is replaced with an encoder-decoder network perhaps because the 1x1x1 convolutional layer can better exploit spatial contextual information. For the local geometry refinement network, we find that global guidance is a vital component as the performance of the local network drops to $0.912$ without it. This is also verified by Fig.~\ref{fig:withoutglobal}, where the final completed point cloud using our model with and without global guidance on two sample models are illustrated.

\begin{table}[]
\centering
\caption{Ablation Study}
\label{table:ablation}
\resizebox{0.47\textwidth}{!}
{
\begin{tabular}{c|l|c}
\hline
Network    & Component           & Performance \\ \hline
           & w/o AUC loss              & 0.904       \\ \cline{2-3}
           & w/o depth images      & 0.877       \\ \cline{2-3}
Global     & w/o BLSTM context modeling      & 0.896       \\ \cline{2-3}
Structure  & w/o BSurf channel               & 0.90        \\ \cline{2-3}
Inference  & BSurf channel only           & 0.836       \\ \cline{2-3}
           & Replace 1x1x1conv            &             \\
           & with encoder-decoder           & 0.818       \\ \cline{2-3}
           & Complete global network         & \bf{0.926}  \\
\hline
Local Geometry & Without global guidance       &    0.912  \\ \cline{2-3}
Refinement     & With global guidance         &    \bf{0.961}\\ \hline
\end{tabular}
}
\end{table}

\section{Conclusion}
We have presented an effective framework for completing partial shapes through 3D CNNs. Our results show that our method significantly improves the performance of existing state-of-the-art methods. We also believe jointly training global and local networks is a promising direction.

\paragraph{Acknowledgements.} The authors would like to thank the reviewers for their constructive comments. Evangelos Kalogerakis acknowledges support from NSF (CHS-1422441, CHS-1617333).
{\small
\bibliographystyle{ieee}
\bibliography{egabib}
}

\onecolumn
\begin{center}
{\Large\bf{High Resolution Shape Completion Using Deep Neural Networks for\\
Global Structure and Local Geometry Inference}\\
{\emph{(Supplementary Material)}}}
\end{center}

\maketitle

\setcounter{section}{0}\section{Visualization of Input}
\begin{figure*}[h]
\centering
\includegraphics[width=0.95\textwidth]{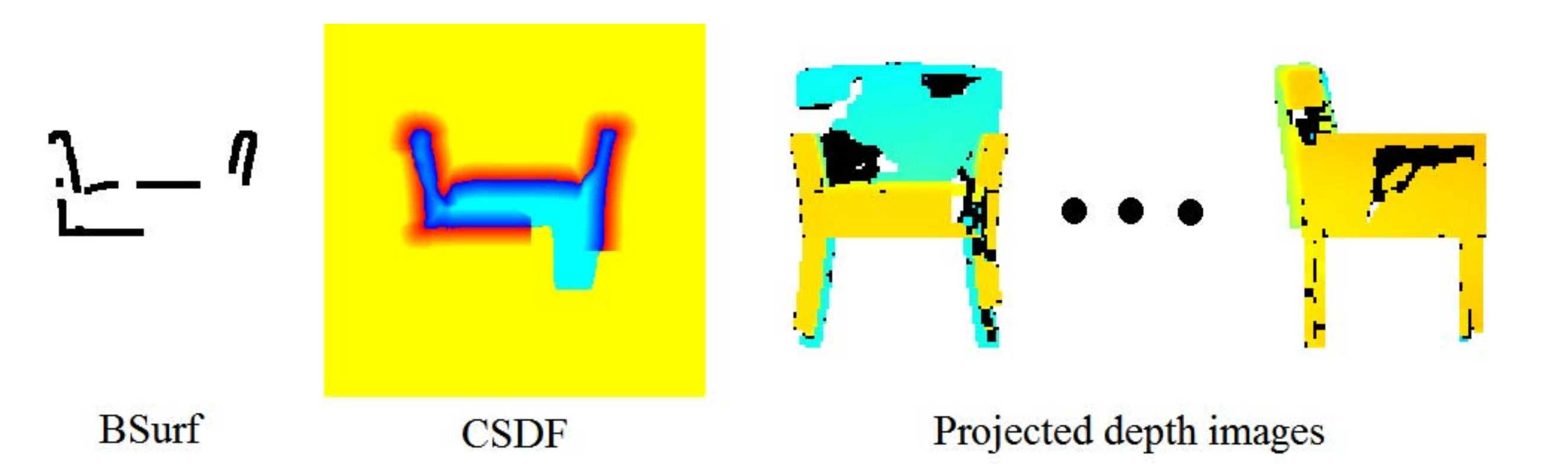}
\caption{Visualization for CSDF+BSurf representation generated from input point cloud and the colored projected depth images.}
\label{fig:inputs}
\end{figure*}

As shown in Fig.~\ref{fig:inputs}, each point cloud is converted into a signed distance field and represented as both a Colored SDF (CSDF) and a binary surface (BSurf). We show an example for each of them by sampling a cross section from a $256^{3}$ voxelized representation. We also show two sampled projected depth images.



\newpage

\section{Sampled Outputs of Global Structure Prediction}

\begin{figure*}[ht]
\centering
\includegraphics[width=0.8\textwidth]{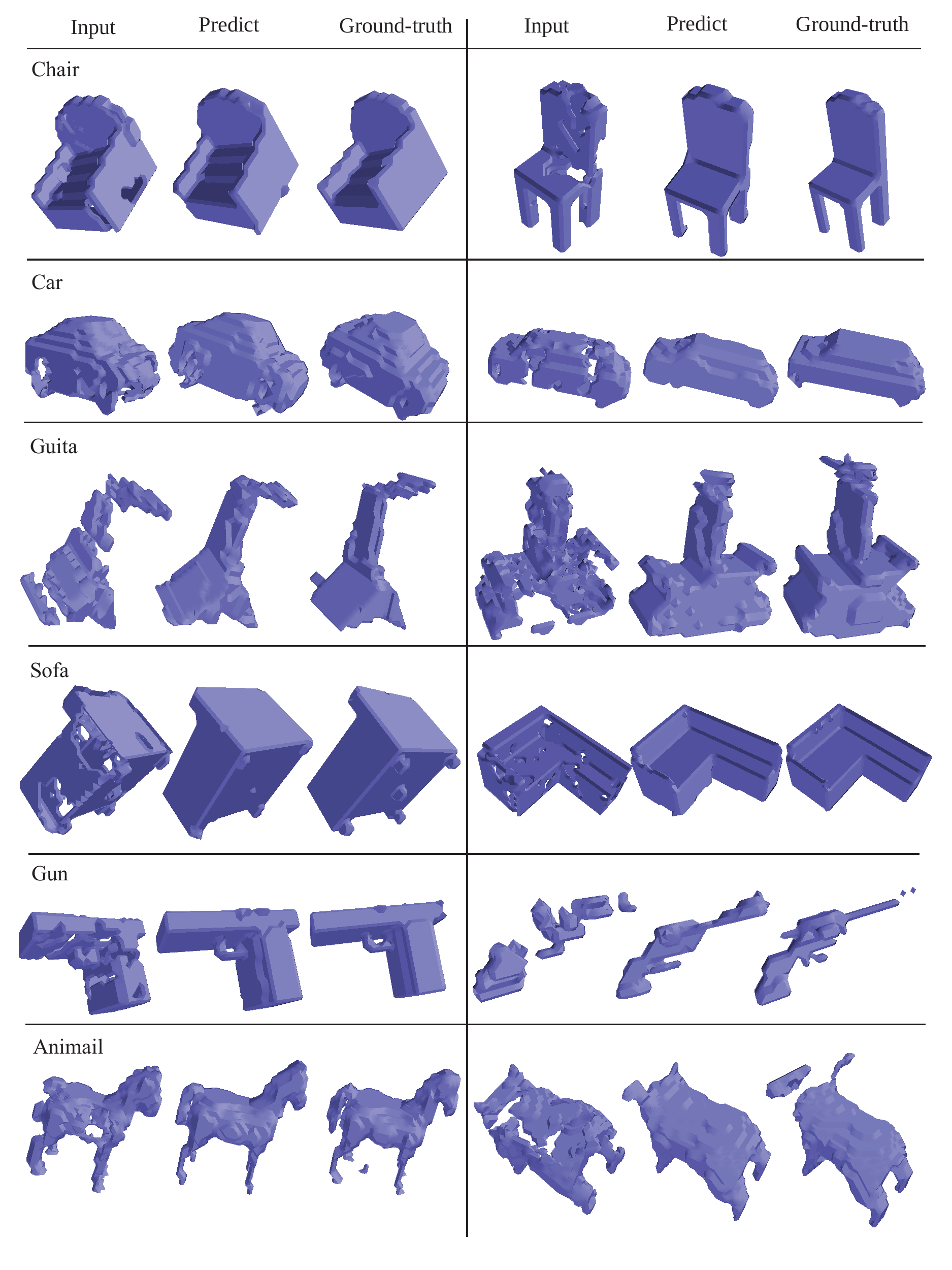}
\caption{Sampled outputs from our global structure inference only. Two models are shown per category. For each model, the input, predicted global structure and ground-truth structure are shown side by side.}
\label{fig:global_outputs}
\end{figure*}

Fig.~\ref{fig:global_outputs} shows a gallery of outputs from our global structure inference network only. We show two results from each object category.

\section{Sampled Outputs from Ablation Study of Global Structure Prediction}
\begin{figure*}[ht]
\centering
\includegraphics[width=0.75\textwidth]{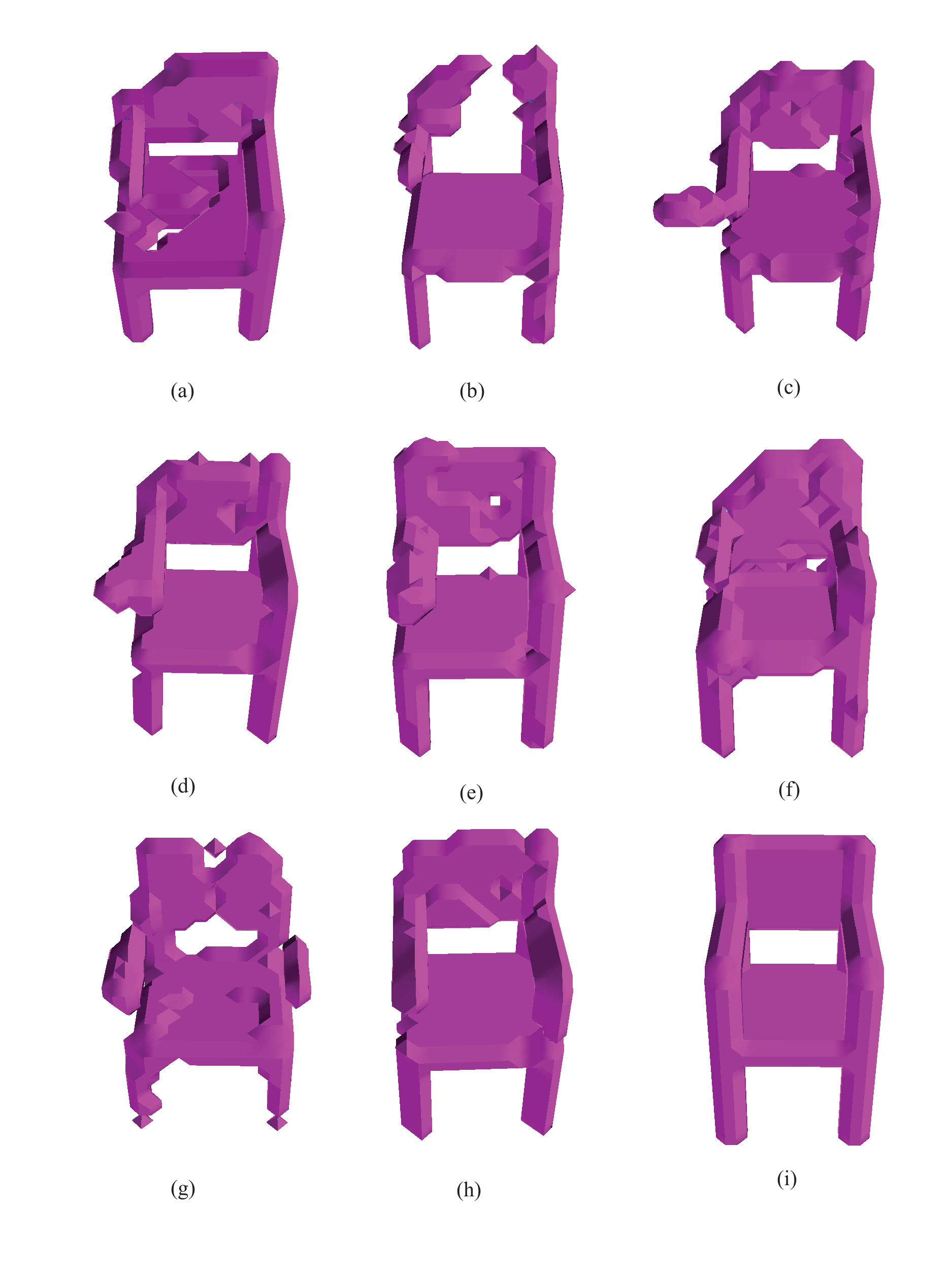}
\caption{Sampled outputs from ablation study of global structure network. (a) is input. (b)-(h) are outputs of w/o AUC loss, w/o depth images, w/o BLSTM context modeling, w/o BSurf channel, BSurf channel only, replacing the 1x1x1 convolution layer with an encoder-decoder and the complete global network, respectively. (i) is the ground-truth global structure.}
\label{fig:global_ablation}
\end{figure*}

For the global structure inference network, we have tested the performance without AUC loss, projected depth images, and global context modeling. In addition, we also tested global models that only take CSDF or BSurf as input to figure out the importance of different input channels. We also tried to replace the 1x1x1 convolutional layer of the global network with an encoder-decoder. For intuitive comparison, Fig.~\ref{fig:global_ablation} shows all the outputs for a sampled chair model.

\section{Comparisons with Poisson Reconstruction}

\begin{figure*}[ht]
\centering
\includegraphics[width=0.75\textwidth]{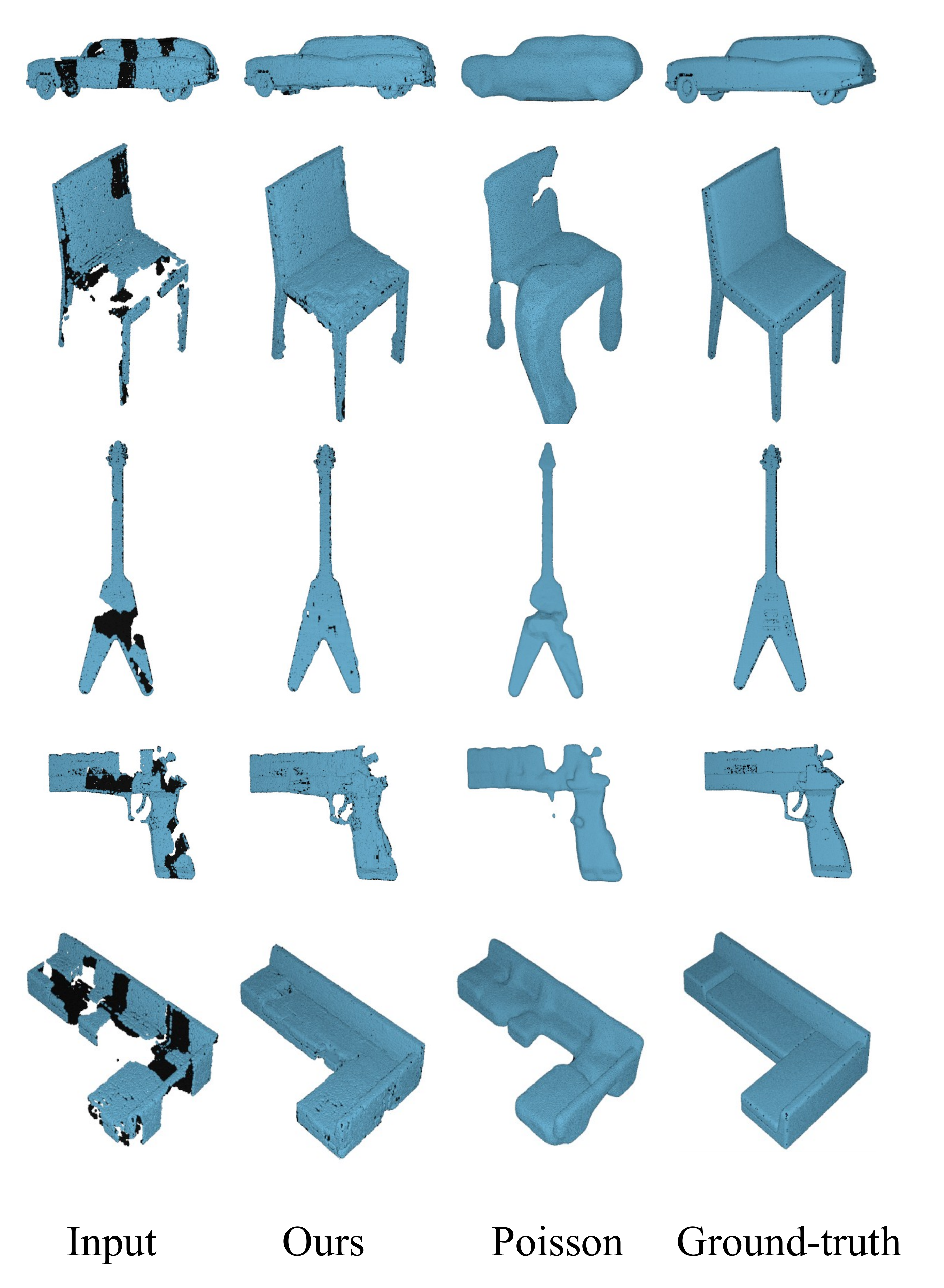}
\caption{Sampled comparison results with Poisson reconstruction method ~\cite{kazhdan2006poisson}.}
\label{fig:comparisonall}
\end{figure*}

To verify the efficiency of our algorithm, we conduct comparisons against Poisson surface reconstruction ~\cite{kazhdan2006poisson}. We show comparison results of 5 sampled models in Fig. ~\ref{fig:comparisonall}, one model per row. For each model, we show the input, the high-resolution completion result of our algorithm, the output of Poisson surface reconstruction ~\cite{kazhdan2006poisson} and the ground-truth side by side.

\newpage

\section{More High-Resolution Shape Completion Results}
\begin{figure*}[h]
\centering
\includegraphics[width=0.9\textwidth]{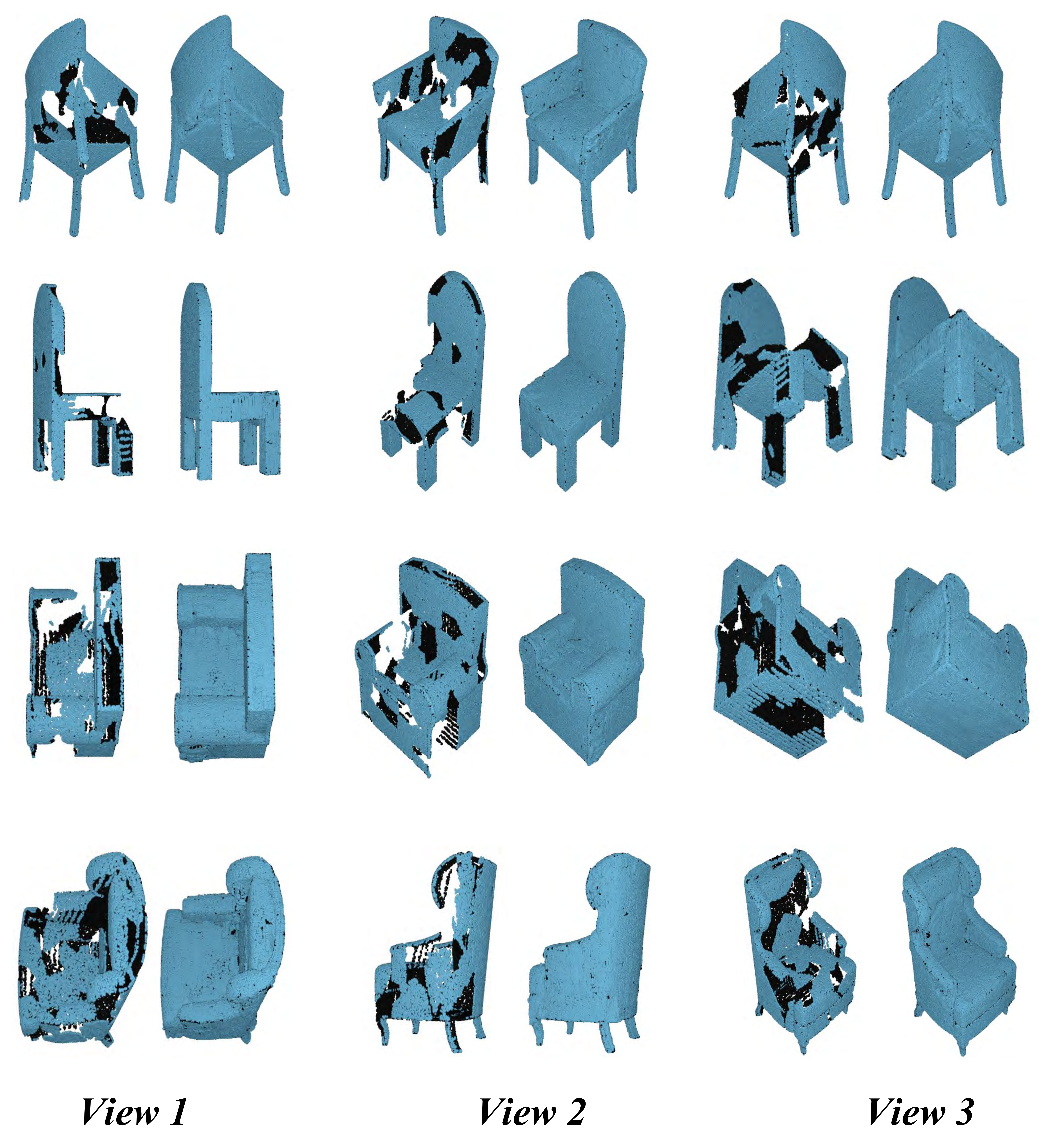}
\caption{Gallery of completion results for the 'chair' category. One model per row. The input and repaired point clouds are
shown side by side from three different views. }
\label{fig:chair}
\end{figure*}

\begin{figure*}[h]
\centering
\includegraphics[width=0.90\textwidth]{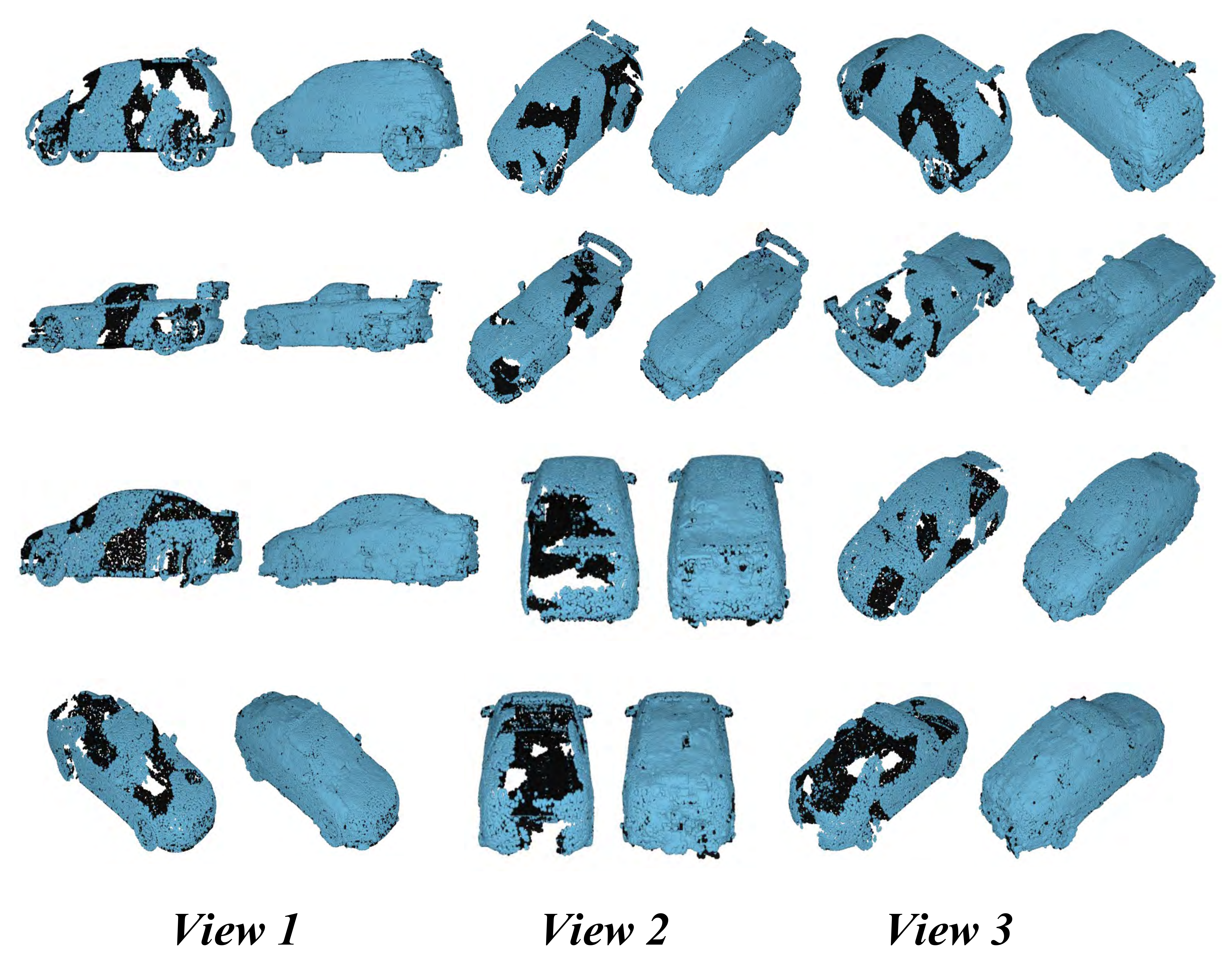}
\caption{Gallery of completion results for the 'car' category. One model per row. The input and repaired point clouds are
shown side by side from three different views.}
\label{fig:car}
\end{figure*}

\begin{figure*}[h]
\centering
\includegraphics[height=0.9\textheight]{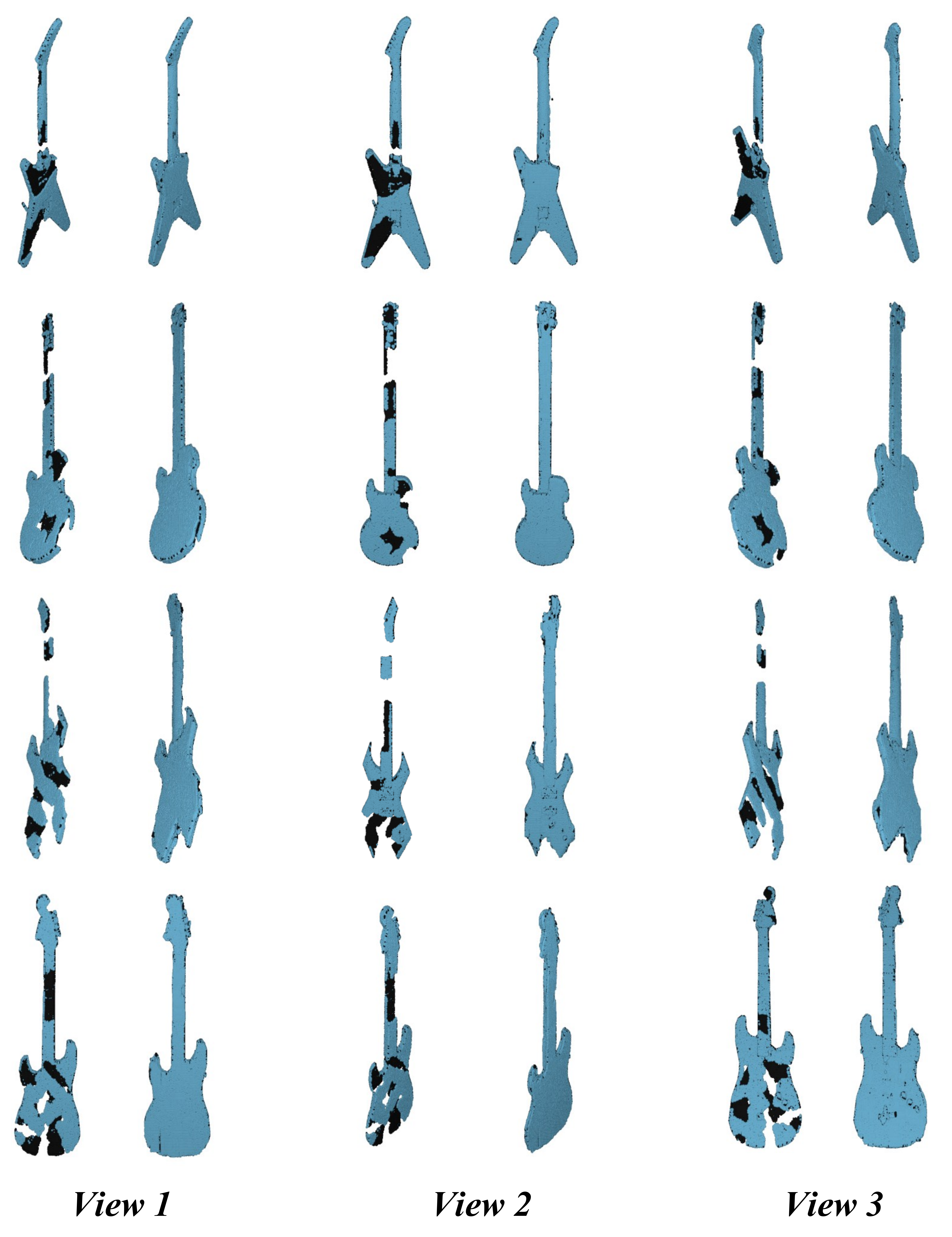}
\caption{Gallery of completion results for the 'guitar' category. One model per row. The input and repaired point clouds are
shown side by side from three different views.}
\label{fig:guita}
\end{figure*}

\begin{figure*}[h]
\centering
\includegraphics[width=0.9\textwidth]{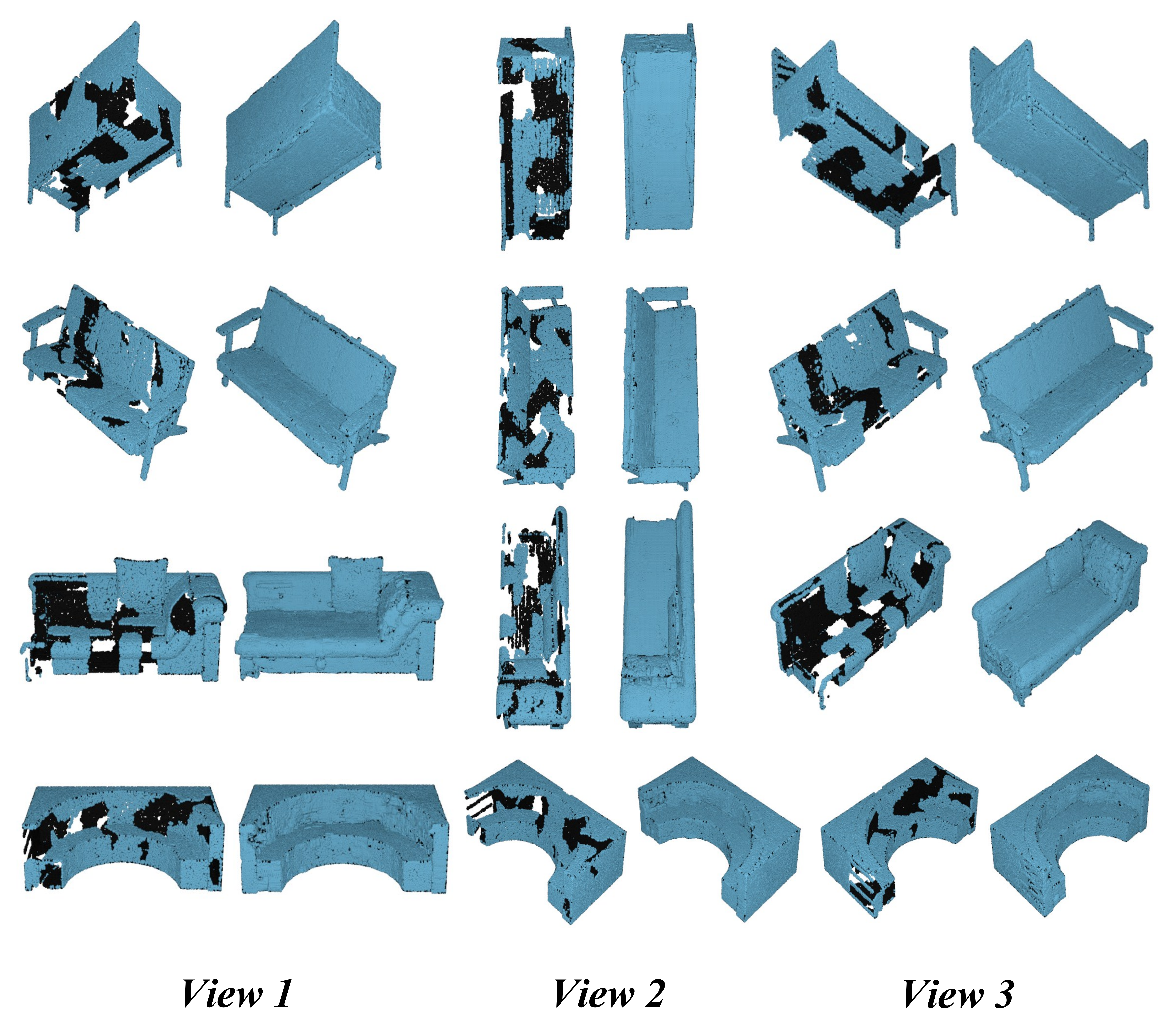}
\caption{Gallery of completion results for the 'sofa' category. One model per row. The input and repaired point clouds are
shown side by side from three different views.}
\label{fig:sofa}
\end{figure*}

\begin{figure*}[h]
\centering
\includegraphics[width=0.9\textwidth]{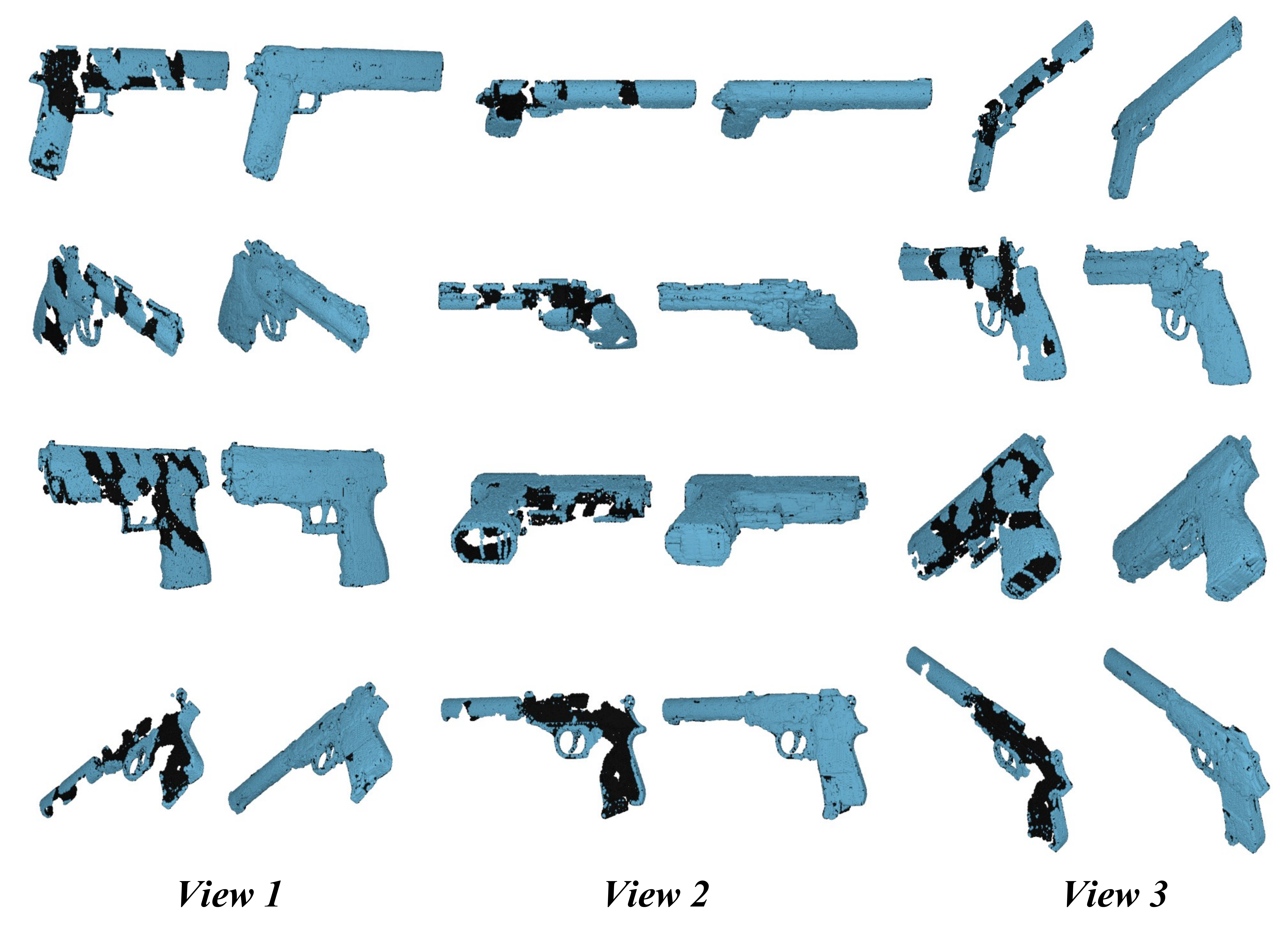}
\caption{Gallery of completion results for the 'gun' category. One model per row. The input and repaired point clouds are
shown side by side from three different views.}
\label{fig:gun}
\end{figure*}

\begin{figure*}[h]
\centering
\includegraphics[width=0.9\textwidth]{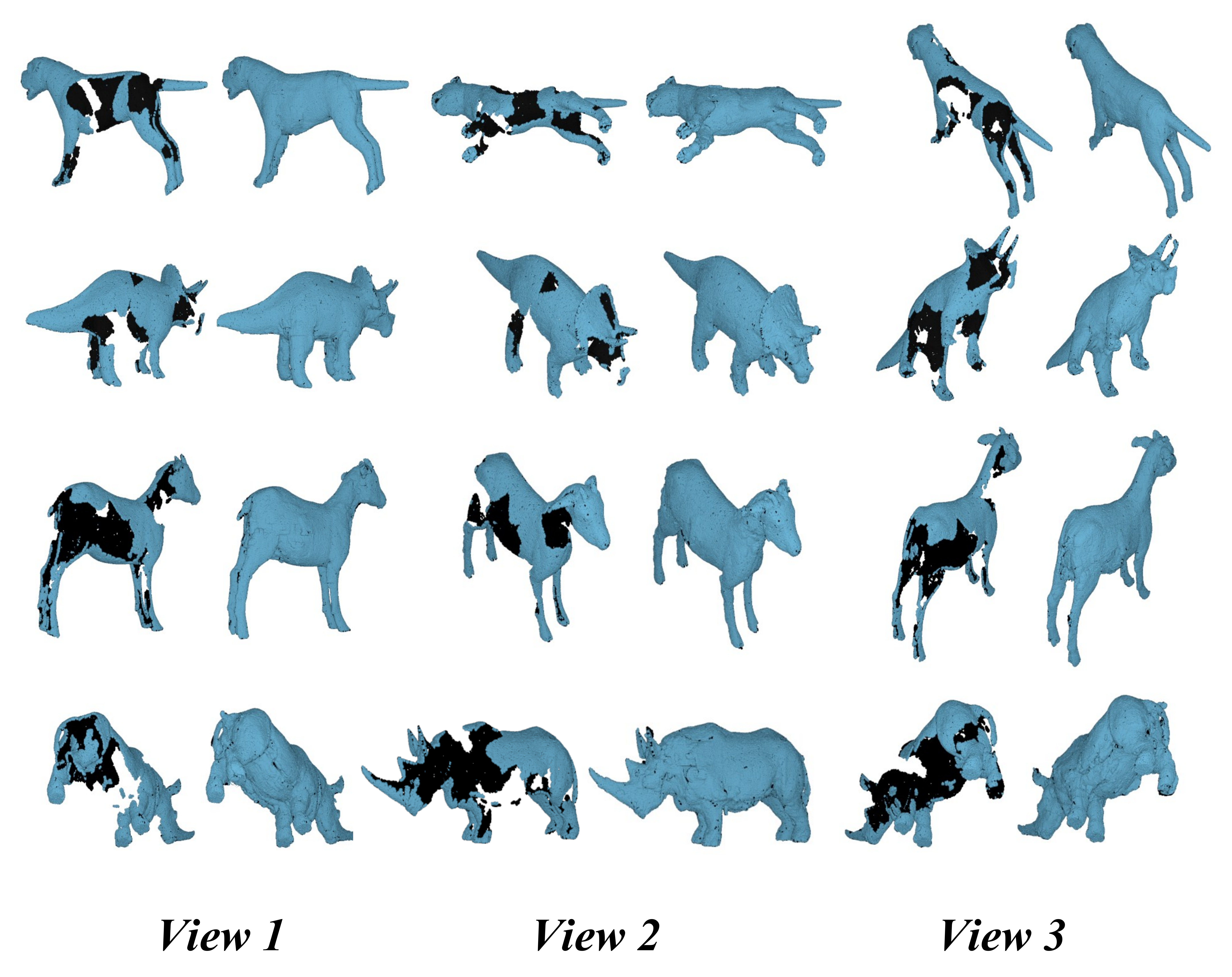}
\caption{Gallery of completion results for  'animal' category. One model per row. The input and repaired point clouds are
shown side by side from three different views.}
\label{fig:animal}
\end{figure*}

Here, we also show more high-resolution shape completion results in Fig.~\ref{fig:chair}-- Fig.~\ref{fig:animal}, one figure per object category.

\newpage
\section{Shape Completion for larger missing regions}
To verify the effectiveness of our algorithm on point clouds with larger missing regions, we re-generated training and testing data and re-trained our joint learning model for the 'chair' category. In this new setting, each partial shape is generated from one or two nearby, randomly selected viewpoints simulating scanning with a real depth sensor, and typically has an entire side missing. The average completeness of an input point cloud is only $55.68\%$. The average completeness and normalized dist of a completed model are $91.67\%$ and 0.00459 respectively. The F1-score of our global network reaches 0.895 and the F1-score of our local refinement network reaches 0.951, both only slightly lower than the ones for point clouds scanned from 3-5 viewpoints. The completion results for two sampled models are shown in Fig.~\ref{fig:singleview}.
\FloatBarrier
{
\begin{figure*}[!h]
\centering
\includegraphics[width=0.9\textwidth]{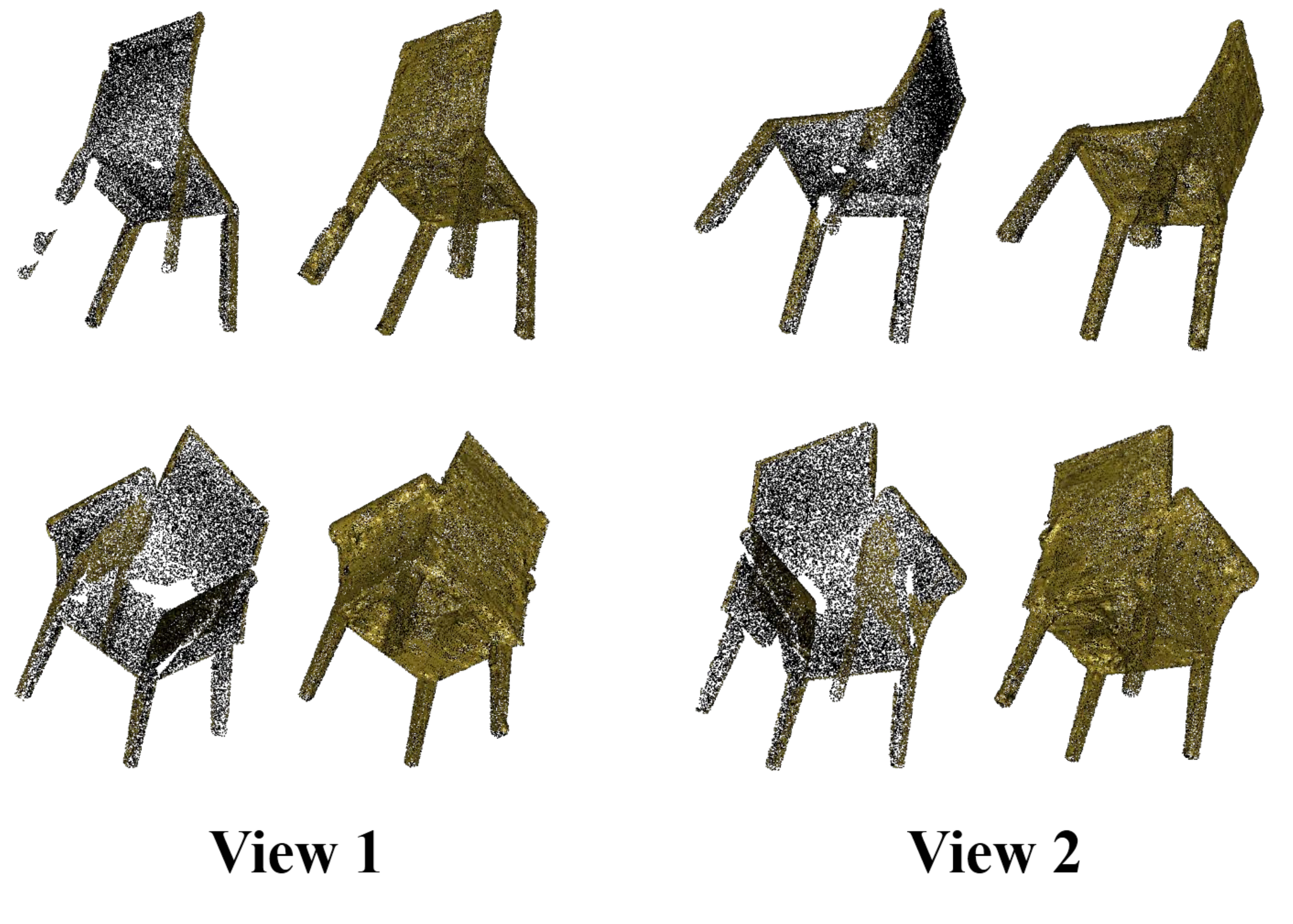}
\caption{Two completion results for two sampled models with large missing regions (one side is entirely missing). One model per row. The input and repaired point clouds are
shown side by side from two different views.}
\label{fig:singleview}
\end{figure*}
%
\end{document}